%% file: main.tex
\documentclass[10pt,twocolumn,letterpaper]{article}

\usepackage[pagenumbers]{cvpr} % To force page numbers, e.g. for an arXiv version
\usepackage{multirow}
\usepackage{caption}
\usepackage{cuted}

\usepackage{array} 
\usepackage{tabularx}
\usepackage{graphicx}
\usepackage{pifont}
\usepackage{siunitx}
\usepackage{setspace}
\definecolor{cvprblue}{rgb}{0.21,0.49,0.74}
\usepackage[pagebackref,breaklinks,colorlinks,allcolors=cvprblue]{hyperref}

\title{
{\includegraphics[height=0.6cm]{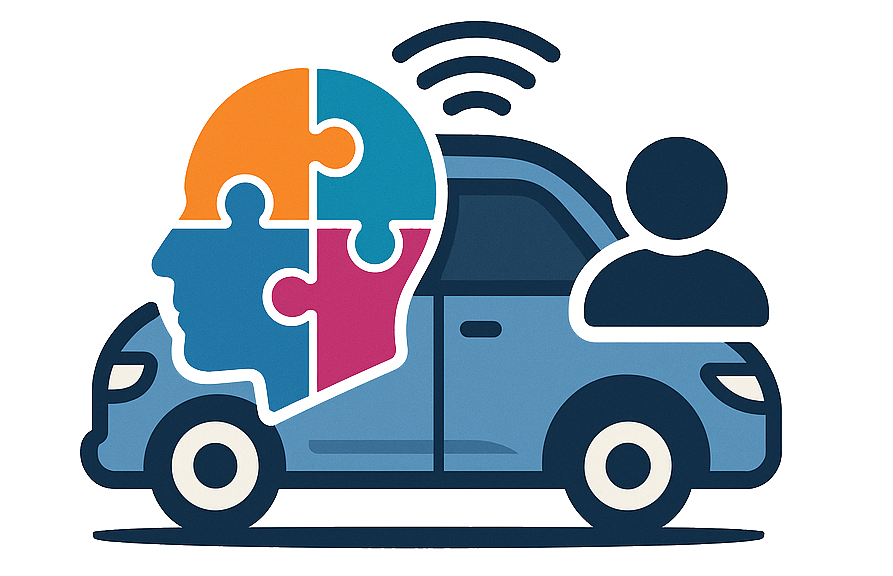}  Drive My Way: Preference Alignment of Vision-Language-Action Model \\ for Personalized Driving}}

\author{
    Zehao Wang$^{1}$,
    Huaide Jiang$^{1}$,
    Shuaiwu Dong$^{1}$,
    Yuping Wang$^{1,2}$,
    Hang Qiu$^{1}$,
    Jiachen Li$^{1,\dagger}$ \\ [0.3em]
    $^{1}$University of California, Riverside \quad
    $^{2}$University of Michigan
}

\begin{document}

\maketitle

\begingroup
\renewcommand{\thefootnote}{}
\footnotetext{
\protect\begin{tabular}[t]{@{}l@{}}
    $^{\dagger}$Corresponding author. \\
    \texttt{zehao.wang1@email.ucr.edu}, \texttt{jiachen.li@ucr.edu} 
    \end{tabular} \\
}
\endgroup

\input{sec/0_abstract}
\input{sec/1_intro}
\input{sec/2_related_work}
\input{sec/3_dataset}
\input{sec/4_formulation}
\input{sec/5_method}
\input{sec/6_experiments}
\input{sec/7_conclusion}

\section*{Acknowledgment}
We gratefully acknowledge Janice Nguyen, Shubham Derhgawen, Jonathan Setiabudi, Alexander Totah, Pranav Gowrishankar, and Pratheek Sunilkumar for their support in data collection.

{
    \small
    \bibliographystyle{unsrtnat}
    \bibliography{main}
}

\appendix \input{sec/X_suppl}
\end{document}

%% file: sec/0_abstract.tex
\begin{abstract}
Human driving behavior is inherently personal, which is shaped by long-term habits and influenced by short-term intentions. Individuals differ in how they accelerate, brake, merge, yield, and overtake across diverse situations. However, existing end-to-end autonomous driving systems either optimize for generic objectives or rely on fixed driving modes, lacking the ability to adapt to individual preferences or interpret natural language intent. 
To address this gap, we propose Drive My Way (DMW), a personalized Vision-Language-Action (VLA) driving framework that aligns with users' long-term driving habits and adapts to real-time user instructions. DMW learns a user embedding from our personalized driving dataset collected across multiple real drivers and conditions the policy on this embedding during planning, while natural language instructions provide additional short-term guidance. Closed-loop evaluation on the Bench2Drive benchmark demonstrates that DMW improves style instruction adaptation, and user studies show that its generated behaviors are recognizable as each driver’s own style, highlighting personalization as a key capability for human-centered autonomous driving. Our data and code are available at \href{https://dmw-cvpr.github.io/}{https://dmw-cvpr.github.io/}.
\end{abstract}

%% file: sec/1_intro.tex
\section{Introduction}
\label{sec:intro}
End-to-end autonomous driving has emerged as a powerful paradigm that directly learns to map raw multi-modal sensor inputs to driving trajectories or control actions~\cite{Jiang_2025_ICCV,gong2026autofocus,wang2025generative,LangeB-RSS-25}. Recent advances in unified architectures and foundation-model-based frameworks have demonstrated impressive performance in open-loop benchmarks through imitation of expert trajectories~\cite{zhao2025diffe2e,li2024adaptive,toyungyernsub2024predicting}. 
However, these systems often optimize for generic safety and efficiency objectives, overlooking the \textit{individuality} and \textit{context-dependent nature} of human driving behaviors. 
In practice, driving is inherently personal; individuals differ in how assertively they accelerate, brake, or overtake depending on the situation, purpose, or emotional state (e.g., commuting, leisure, or emergency travel). 
Existing autonomous driving systems typically provide only a few preset modes (e.g., ``sport'', ``comfort'', or ``eco'') or manual parameter adjustments, which fail to capture subtle and evolving passenger preferences~\cite{kou2025padriver, nakanoya2021personalized,li2024interactive}. 
More importantly, such rigid presets cannot interpret intuitive, natural language instructions, such as ``I’m tired'' or ``I’m late for work''. Therefore, achieving \textit{long-term, context-adaptive personalization} is crucial for enhancing user trust, comfort, and satisfaction~\cite{hao2026styledrive,ghosh2026reducing}.

\begin{figure}[!t]
  \centering
   \includegraphics[width=\linewidth]{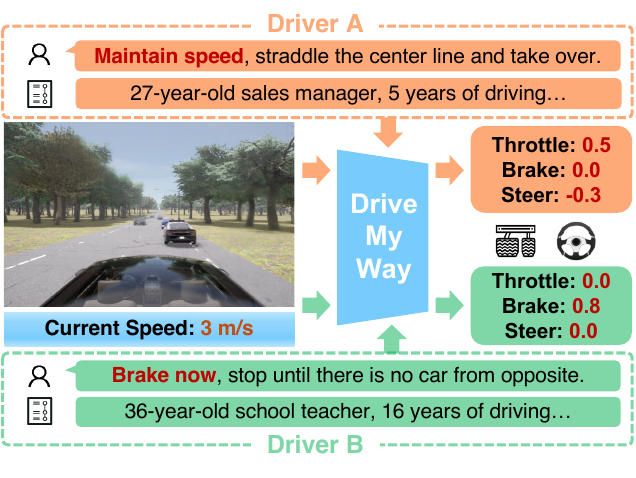}
   \vspace{-0.5cm}
   \caption{Drive My Way (DMW) achieves end-to-end personalized driving via both long-term preference alignment and short-term style instruction adaptation.}
   \vspace{-0.3cm}
   \label{fig:teaser}
\end{figure}

Existing research on personalized autonomous driving mainly falls into two categories. 
First, data-driven methods extract predefined driving styles from human demonstrations and learn style-conditioned policies using behavior cloning or inverse reinforcement learning (IRL)~\cite{hao2026styledrive, jiang2025irl}. While effective at mimicking human-like styles, these approaches require diverse large-scale datasets and scale poorly to a growing population of users with heterogeneous preferences. Surmann et al.~\cite{surmann2025multi} mitigate this limitation through multi-objective reinforcement learning, enabling runtime adjustment of preference weights. However, such methods cannot handle real-time human interaction via natural language or adapt to user instructions on the fly. 
Second, language-driven approaches leverage large language models (LLMs) for instruction-based personalization~\cite{LaMPilot}. 
For example, Talk2Drive~\cite{cui2024personalized} employs GPT-4 to interpret passenger commands and generate high-level driving decisions, reducing takeover rates in simplified intersection and parking scenarios involving limited traffic participants. 
Despite promising results, existing language-based methods remain constrained to simple, low-interaction settings. They have not been systematically investigated in complex, dynamic driving contexts, such as merging, overtaking, or emergency braking, where trade-offs among safety, comfort, and efficiency must adapt to dynamic scenes. 
Fully leveraging the reasoning and generalization capabilities of foundation models for such context-aware adaptations remains largely unexplored.
Additionally, current personalization methods based on LLMs~\cite{cui2024personalized} or vision-language models (VLMs)~\cite{cui2024board} do not consider a driver’s \textit{long-term habits} or cumulative experiences, which are diverse and continuously evolving. 

To address these issues, we propose DMW, a novel Vision-Language-Action (VLA) framework that integrates visual observations, contextual user profiles, and real-time natural language instructions to generate adaptive, personalized driving actions, as shown in Fig.~\ref{fig:teaser}. DMW learns user embeddings that encode long-term driving behaviors from a newly collected personalized driving dataset comprising profiles, trajectories, and privileged information from thirty drivers across twenty realistic scenarios in CARLA~\cite{Dosovitskiy17}. To enable human interaction and real-time adaptation, we further apply reinforcement fine-tuning to the driving policy using rewards derived from safety, comfort, and efficiency objectives, whose weights are dynamically adjusted. Extensive closed-loop experiments on Bench2Drive~\cite{jia2024bench2drive} show that DMW effectively adapts to different style instructions while maintaining safety. Further user studies demonstrate that its driving behavior aligns with individual user preferences and expresses distinct driving behaviors.

Our main contributions are summarized as follows:
\begin{itemize}
\item We propose DMW, a novel personalized end-to-end autonomous driving framework that integrates contextual user embeddings to align policy behaviors with individual driving preferences. DMW also enables real-time human interaction and adaptability through reinforcement fine-tuning conditioned on natural language instructions.
\item We construct the first multi-modal personalized driving dataset (PDD) collected from thirty real drivers across diverse traffic scenarios in CARLA, which provides a valuable dataset for developing and evaluating human-centered driving models.
\item We conduct extensive evaluations on the Bench2Drive benchmark~\cite{jia2024bench2drive}, complemented by personalization metrics and user studies, which demonstrate the effectiveness of DMW in adapting its driving behavior to individual user preferences while maintaining a balanced trade-off among safety, efficiency, and comfort.
\end{itemize}

%% file: sec/2_related_work.tex
\section{Related Work}
\label{sec:related_work}

\noindent\textbf{Foundation Models for Autonomous Driving.} 
Existing end-to-end autonomous driving models \cite{jia2023driveadapter, jia2023thinktwice, zhou2025autovla} still struggle in closed-loop benchmarks \cite{carlaleaderboard,wang2025uniocc} due to reliance on imitation learning and limited generalization \cite{chi2025impromptu, LinModelBasedPA}. 
Foundation models (LLMs, VLMs) have recently emerged \citep{vlm3r, 11370877, gupta2026scale, zhou2024embodied, zhang2026commcp, yan2025rdd, zhang2025lamma, chakraborty2025heal, nag2025conformal, TrajEvo} and existing work (e.g., DriveVLM \citep{tian2025drivevlm}, GPT-Driver \citep{mao2023gpt}, AlphaDrive \citep{jiang2025alphadrive}) leverages their world knowledge and semantic reasoning to interpret complex traffic scenarios, producing high-level decisions and language rationales for their actions. In parallel, VLA models have demonstrated the ability to directly translate raw sensory inputs and linguistic instructions into fine-grained actions \citep{Fu_2025_ICCV, zeng2025futuresightdrive} or low-level control signals through action tokens \citep{zhou2025autovla} or action decoders \citep{renz2025simlingo}. While these approaches have made substantial progress toward language-conditioned driving, their capacity for personalization and adaptation to individual driving preferences remains largely underexplored.

\noindent\textbf{Personalization in Autonomous Driving.} The pursuit of personalized autonomous driving remains an active area of research, with the key challenge lying in aligning individual driving preferences. Recent studies like \citep{hao2026styledrive} learn style-conditioned driving policies, but overlook fine-grained individual differences. MAVERIC \citep{schrum2024maveric} explores the learning of a latent space of diverse and socially-aware driving behaviors, offering greater flexibility compared to predefined style categories. Meanwhile, LLMs have enabled more intuitive human-vehicle interaction \citep{kou2025padriver, cui2024board, han2024words, LaMPilot}, allowing systems to respond to personalized commands in real time. 
Recent works~\cite{cui2024personalized} leverage memory modules based on retrieval-augmented generation (RAG) to personalize decisions by retrieving user-specific context.
However, these approaches rely on explicit human feedback and accurate retrieval, producing contexts that are largely descriptive rather than behavioral and fail to capture implicit, long-term driving tendencies.
Moreover, most of them are evaluated only in open-loop settings, making it difficult to assess how well personalization transfers to closed-loop decision making and influences real-time driving behavior.

%% file: sec/3_dataset.tex
\section{Personalized Driving Dataset}
\label{sec:personalized_driving_dataset}

\begin{figure}[!t]
  \centering
   \includegraphics[width=\linewidth]{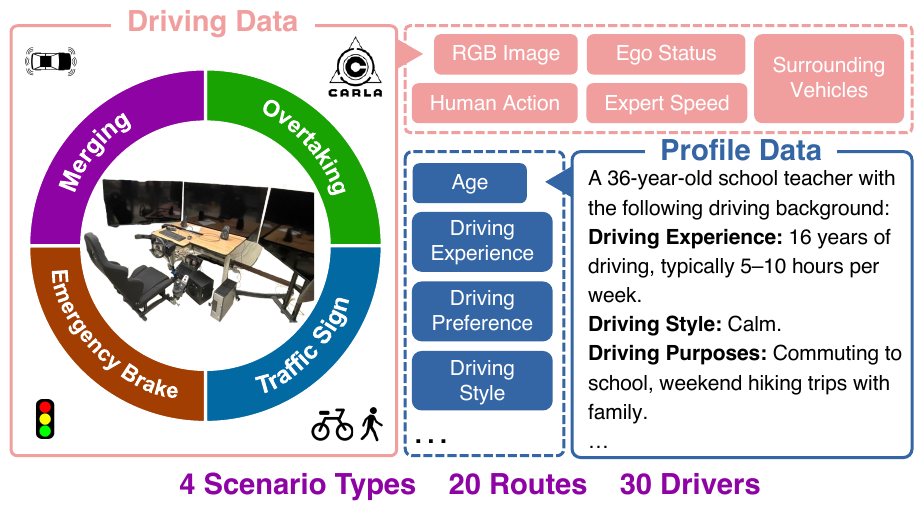}
   \vspace{-0.5cm}
   \caption{An overview of the Personal Driving Dataset, which consists of the driving data and structured driver profile data.}
   \vspace{-0.5cm}
   \label{fig:dataset}
\end{figure}

To learn and evaluate personalized driving policies, we construct a new \textbf{Personalized Driving Dataset (PDD)}, as illustrated in \cref{fig:dataset}. 
PDD captures the driving environments and the personal context of human drivers across diverse, highly interactive, safety-critical driving situations.
Unlike prior datasets that rely on synthetic instruction-action pairs with limited behavioral variation~\cite{qin2025investigating} or those that infer style labels retrospectively using coarse VLM-based reasoning or heuristic labeling~\cite{hao2026styledrive}, our dataset records human driving behaviors during complex real-time interactions, which allows us to study how different individuals make decisions when faced with similar situations.

We recruit thirty drivers with diverse backgrounds and levels of driving experience. Before data collection, each participant completes a structured questionnaire covering demographic information, driving history, and typical driving purposes such as commuting or leisure travel. 
These profiles provide a semantic context that connects personal background with long-term driving habits. Each participant then performs a standardized set of twenty driving scenarios in the CARLA, spanning four scenario types, including overtaking, merging into traffic, handling intersections, and navigating pedestrian crossings or encountering vehicle cut-ins. 
All drivers operate the vehicle using a Logitech G-series steering wheel and pedal setup to enable naturalistic control inputs. 
PDD records ego-vehicle motion states, scene perception of surrounding vehicles, pedestrians, cyclists, and roadside hazards, as well as traffic context such as signal states, stop signs, route geometry, and speed limits. 
To support comparison across drivers and scenarios, we record an expert target speed using PDM-Lite~\cite{sima2024drivelm}; the deviation between human-driven speed and this target serves as a dense descriptor of behavioral style under varying conditions.
Overall, PDD offers a rich and behavior-sensitive foundation for modeling personalized driving, which supports preference-conditioned policy learning.

%% file: sec/4_formulation.tex
\section{Problem Formulation}
\label{sec:formulation}

We aim to learn a personalized driving policy that aligns with a driver's long-term driving behavior while adapting to real-time preference instructions. 
Let $\mathcal{M}=\{1,\ldots,M\}$ denote the set of drivers, where $M$ is the number of drivers. For each driver $m \in \mathcal{M}$, the collected driving dataset is $\mathcal{D}^m = \big\{(s_{t}^{m}, a_{t}^{m})\big\}_{t=1}^{T^m} $, where $s_{t}^{m}$ and $a_{t}^{m}$ denote the environmental state and the executed action of driver $m$ at time step $t$, and $T^m$ is the trajectory length. 
The user profile obtained is $P^m$, which summarizes the driver’s background, habits, and driving experience. The complete dataset across all drivers is $\mathcal{D} = \bigcup_{m \in \mathcal{M}} \mathcal{D}^m.$

We formalize personalized driving as a Markov Decision Process (MDP), defined by the tuple $(\mathcal{S}, \mathcal{A}, \mathcal{O}, \mathcal{T}, \mathcal{R}, \gamma)$. $\mathcal{S}$ represents the state space, which includes environmental information. $\mathcal{A}\subset\mathbb{R}^3$ is the continuous action space consisting of throttle, brake, and steering. At time $t$, the agent observes $o_t = (\mathbf{I}_t, \mathbf{q}_t, I_t, g_t, P^m) \in \mathcal{O}$, where $\mathbf{I}_t \in \mathbb{R}^{H \times W \times 3}$ is the front-view image, $\mathbf{q}_t$ is the ego-vehicle state, $I_t$ is a language instruction expressing short-term preference (e.g., ``I’m in a rush''), $g_t$ is the navigation target (e.g., route waypoints), and $P^m$ identifies the context of the active driver.
$\mathcal{T}$ denotes the transition in the environment. $\mathcal{R}$ is the reward function that evaluates the action. $\gamma$ is the discount factor.

The objective is to learn a driving policy $\pi_\theta(a_t \mid o_t)$ that maximizes the expected cumulative discounted reward: $\max_{\theta}\mathbb{E}_{\pi_\theta} \left[ \sum_{t=0}^{T} \gamma^{t}\mathcal{R}(s_t,a_t) \right].$ To achieve personalization, the policy must infer the driver’s long-term behavioral tendencies from their historical driving data $\mathcal{D}^m$ and relate these behaviors to their profile $P^m$. During execution, the policy adapts its decisions based on the specified user embedding $z_p^m$, which is encoded from the driver’s profile $P^m$, and the current instruction $I_t$, ensuring safe and effective driving behavior that reflects each driver’s characteristic style and situational preferences.

%% file: sec/5_method.tex
\section{Method}
\label{sec:method}

\begin{figure*}[!t]
  \centering
   \includegraphics[width=\textwidth]{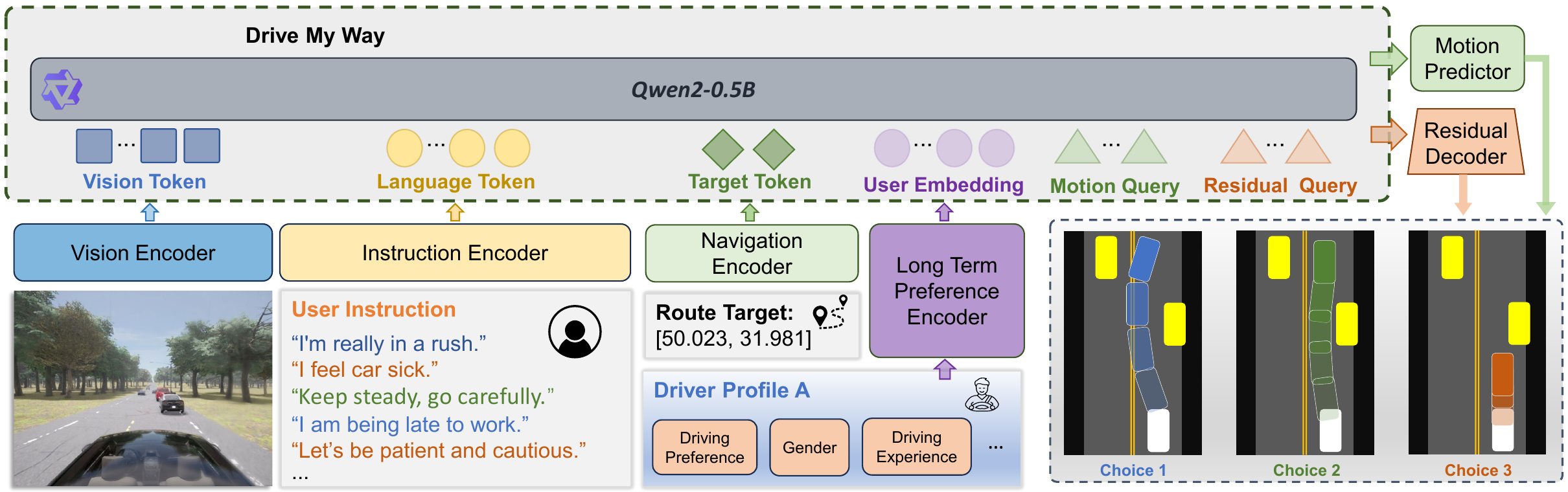}
    \vspace{-0.5cm}
   \caption{An overview of the DMW framework with a pretrained VLA backbone. The model takes in front-view camera images, instructions, route target points, and user profile as inputs, while the motion predictor outputs route and speed waypoints, which derive the base action (throttle, steer angle). The residual decoder outputs a discrete residual applied to the base to produce the final personalized action.}
   \vspace{-0.3cm}
   \label{fig:arch}
\end{figure*}

We focus on personalized adaptation of driving policies to accommodate different users and their short-term driving preferences. The overall diagram of DMW is provided in \cref{fig:arch}. 
Given the camera observations and navigation goals, the model fuses the driver’s long-term preference with user instructions to produce adaptive, personalized actions.

\subsection{VLA Backbone}

We employ SimLingo~\cite{renz2025simlingo} as our VLA backbone due to its ability to perform language grounding and planning. SimLingo is built on InternVL2-1B~\cite{chen2024far, chen2024internvl}, which integrates the InternViT-300M-448px vision encoder with the Qwen2-0.5B language model, offering computational efficiency compared to larger alternatives~\cite{yang2024qwen2}. 
Supervised by privileged expert demonstrations from PDM-Lite ~\cite{sima2024drivelm} and auxiliary reasoning signals, the backbone predicts temporal and geometric waypoints, which are then converted into target speed and steering commands.

\subsection{Personalization via Reinforcement Fine-tuning}

Despite the strong planning capability of the VLA backbone, imitation learning alone tends to capture generalized behaviors rather than user-specific driving styles. 
We employ the Group Relative Policy Optimization (GRPO) to further finetune the personalization ability \cite{shao2024deepseekmath}. 
By sampling a group of outputs and computing normalized group advantages, GRPO  inherently promotes policy specialization. To generate a rich set of candidate actions for the GRPO policy update, we introduce a residual decoder.
As shown in \cref{fig:arch}, learnable \emph{residual query} tokens are integrated into the language model along with vision, language, navigation, user embedding, and motion query tokens. The resulting residual features are decoded by an MLP followed by a categorical action head to produce two discrete residual adjustments: (i) speed change and (ii) steering change. 

The combined action is then passed through a PID controller to produce the final $a_t = a_t^{\text{base}} + a_t^\Delta$, where $a_t^{\text{base}}$ is derived from the predicted waypoints and $a_t^\Delta$ is the personalized residual. This design allows the policy to preserve safe planning while adapting expressively to different drivers and situational intentions among multiple feasible planning trajectories (i.e., different choices of action).

\subsection{Long-term Preference Learning and Alignment}

\begin{figure}[!t]
  \centering
   \includegraphics[width=\linewidth]{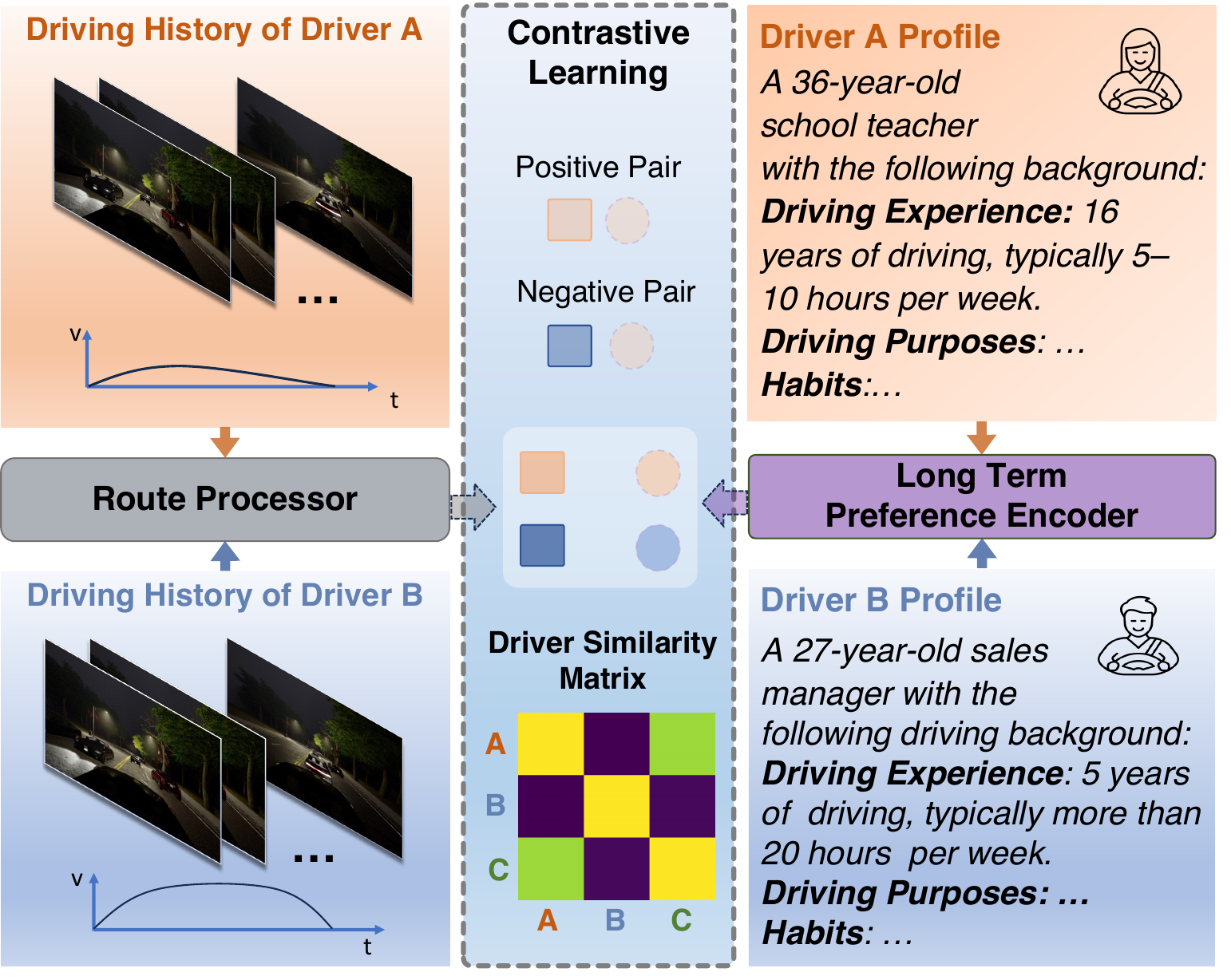}
   \vspace{-0.5cm}
   \caption{The contrastive learning mechanism on the long-term preference encoder and route processor.}
   \vspace{-0.3cm}
   \label{fig:pref_encoder}
\end{figure}

\textbf{User Embedding Learning.}
To model the underlying driving style of different users and align VLA with personalized preferences, we introduce a long-term preference encoder that learns user embeddings from profile. To relate the semantic profile context with corresponding driving behavior, we adopt a contrastive learning mechanism~\cite{oord2018representation} to learn a shared latent space $\mathcal{Z}$.
As illustrated in \cref{fig:pref_encoder}, the long-term preference encoder $f_p(\cdot)$ takes a profile $P^m$ of driver $m$ from the total set of drivers $M$ and outputs a user embedding $z_p^m \in \mathcal{Z}$. $f_p(\cdot)$ includes a DeBERTaV3~\cite{he2023debertav} text processor followed by a projection head.
The route processor $f_b(\cdot)$ is a temporal encoder with multi-head self-attention that processes a past trajectory window of length $k$ at the current timestep $t$, 
$\mathcal{\xi}_{t}^m = \{(\mathbf{I}_{t-k:t}^m, \mathbf{q}_{t-k:t}^m, a_{t-k:t}^m)\}$, 
which consists of sequential front-view camera images $\mathbf{I}_{t-k:t}^m$, ego-vehicle states $\mathbf{q}_{t-k:t}^m$ and driver's actions $a_{t-k:t}^m$ from $D^m$ in PDD. This outputs a behavior embedding $z_{b,t}^m \in \mathcal{Z}$ that summarizes the driver’s actual driving tendencies.
To align the embeddings from the two encoders, we employ the InfoNCE \cite{oord2018representation} contrastive objective that encourages $z_p^m$ and $z_{b,t}^m$ from the same driver to be close while pushing embeddings from different drivers apart:
\begin{equation}
\mathcal{L}_{t}^m = -\log 
\frac{\exp\left(\mathrm{sim}(z_p^m, z_{b,t}^m)/\tau\right)}
{\sum_{j=1}^{M} \exp\left(\mathrm{sim}(z_p^j, z_{b,t}^m)/\tau\right)},
\end{equation}
where $z_p^j$ is the user embedding of all other drivers, $\mathrm{sim}(\cdot, \cdot)$ denotes the cosine similarity and $\tau$ is the temperature.

\noindent\textbf{Preference Alignment.}
After obtaining the user embedding $z_p^m$, we condition the VLA policy on this embedding and further adapt it through reinforcement fine-tuning. The user embedding acts as a latent personalization prior, guiding the policy toward decision patterns consistent with the driver’s characteristic style. 

To align the policy with individual driving preferences while enhancing behavioral diversity, for each target driver $m$, we augment trajectories by conditioning the user embedding on both their own and another driver’s data $u \in \mathcal{M},\, u \neq m$. We want driver $m$ and driver $u$ to behave differently in order to effectively distinguish different drivers, so for a given driver $m$, we always select $u$ who has the least similarity with $m$ regarding user embedding: $u=\arg \min_{x\in \mathcal{M}} \mathrm{sim}(z_p^m, z_p^x)$. The augmented action $\tilde{a}_t^{m}$ is estimated by scaling the original human action $a_t^{m}$ according to the ratio of their route-level action statistics, which capture each driver’s action variability. Let the route-level average action of driver $m$ and $u$ be $\bar{a}^m$ and $\bar{a}^u$ respectively, then we calculate the augmented action as: $\tilde{a}_t^{m}=\frac{\bar{a}^m}{\bar{a}^u} \cdot a_t^m$. 
The reward is then formulated as a behavioral similarity between the model’s sampled action and the reference or augmented action. When the policy is conditioned on the target profile $P^m$: $\mathcal{R}(s_t^m, a_t) = d(a_t, a_t^m)$. When conditioned on $P^u$: $\mathcal{R}(s_t^m, a_t) = d(a_t, \tilde{a}_t^m)$, where $d(\cdot)$ measures similarity in action space, encouraging the policy to adapt decisions to each user embedding while learning a smooth manifold of diverse driving styles.

\subsection{Human-Vehicle Personalized Interaction}

\textbf{Style Instruction.} 
In complex and dynamically evolving traffic scenarios, effective personalization requires the autonomous agent to infer not only the long-term driving preference but also the context-adaptive preferences that vary across situations.
To capture these nuances, we construct a style instruction set spanning twenty distinct driving scenarios. 
These instructions incorporate style intent and scene-specific semantics. Each scenario includes nine stylistic instructions that cover three driving styles expressed at three levels of directness, following the degrees of implicitness defined in prior work~\cite{cui2024personalized}. 
This design enables the model to interpret both explicit commands and subtle linguistic cues that reflect real-time user intent across diverse contexts, thereby achieving context-aware adaptation.

\begin{table*}[!t]
\centering
\caption{\textbf{Bench2Drive closed-loop driving metrics with different style instructions.} 
We compare SimLingo and StyleDrive under different style instructions with our policy fine-tuning with fixed rewards weights ({DMW-Vanilla}) and style-aware rewards weights ({\textbf{DMW}}).}
\vspace{-0.2cm}
\small
{\renewcommand{\arraystretch}{0.9} % adjust row spacin
{\begin{tabular}{l|c|>{\centering\arraybackslash}ccccccccc}
\toprule
\textbf{Method} & Style & DS & SR & Efficiency & Comfort & Speed & Acceleration & LC & Headway & TT \\ 
\midrule
SimLingo~\cite{renz2025simlingo} & Aggressive & 78.56 & 65.83 & 247.60 & 18.61 & 7.66 & 5.39 & 0.75 & 25.99 & 25.35\\ 
& Neutral & 78.15 & 65.85 & 241.44 & 24.67 & 7.37 & 5.22 & 0.75 & 27.81 & 31.41 \\
& Conservative  & 78.18 & 65.56 & 238.77 & 26.99 & 7.21 & 5.29 & 0.70 & 29.12 & 33.02 \\
\midrule
StyleDrive~\cite{hao2026styledrive}
& Aggressive & 75.68 & 60.89 & 256.71 & 16.79 & 7.23 & 5.59 & 0.74 & 24.95 & 27.76 \\
& Neutral & 76.26 & 62.13 & 249.07 & 21.35 & 6.98 & 5.43 & 0.66 & 23.62 & 29.12 \\
& Conservative & 77.02 & 61.96 & 242.18 & 23.67 & 6.82 & 5.39 & 0.70 & 27.19 & 29.98 \\
\midrule
DMW-Vanilla & Aggressive & 82.19 & 70.97 & 253.10 & 15.86 & \textbf{7.86} & 5.29 & \textbf{0.78} & 26.46 & 19.69 \\
& Neutral & 81.96 & 70.63 & 247.77 & 19.21 & 7.66 & 5.17 & 0.77 & 26.63 & 23.16 \\
& Conservative & 81.48 & 71.05 & 246.80 & 21.87 & 7.75 & 5.37 & 0.75 & 26.90 & 22.51    \\
\midrule
\textbf{DMW} & Aggressive & 79.50 & 67.36 &\textbf{281.56} & 21.62 & 7.72 & \textbf{6.01} & 0.70 & 26.37 & 26.93 \\
& Neutral & 82.03 & 70.95 & 244.98 & 28.67 & 6.34 & 5.43 & 0.61 & 27.60 & 40.75 \\
& Conservative & \textbf{82.72} & \textbf{71.56} & 237.06 & \textbf{34.62}& 6.18 & 5.26 & 0.60 & \textbf{30.05}& \textbf{47.38}\\
\bottomrule
\end{tabular}}}
\label{tab:bench2drive}
\end{table*}

\noindent\textbf{Style-Aware Reward Adaptation.}
To translate user intents into safe and personalized objectives, we design a reward function that integrates driving performance metrics with instruction-dependent style alignment. The overall driving reward is formulated as a weighted combination of safety, efficiency, and comfort terms:
\begin{equation}
\mathcal{R}(s_t, a_t) = w_s \cdot R_\text{safety} + w_e \cdot R_{\text{efficiency}} + w_c \cdot R_{\text{comfort}},
\label{eq:reward}
\end{equation}
where $w_s$, $w_e$, and $w_c$ are importance weights, and $R_{\text{safety}}$, $R_{\text{efficiency}}$, and $R_{\text{comfort}}$ denote the corresponding components.
The safety reward penalizes risky interactions based on Time-to-Collision (TTC):
$R_{\text{safety}} = \mathbb{I}_\text{safety}\big(\text{TTC}_t \geq \beta_\text{safety}\big)$,
where $\mathbb{I}_\text{safety}$ is the binary indicator enforces a minimum instruction-dependent safety threshold $\beta_\text{safety}$. 
The efficiency reward encourages maintaining a speed consistent with the desired style:
$R_{\text{efficiency}} = \exp(-\alpha \cdot |v_t - v_{\text{pref}}|)$,
where $v_t$ is the actual speed derived from action $a_t$, $v_{\text{pref}}$ is the instruction-dependent preferred speed, and $\alpha$ is a penalty coefficient controlling sensitivity to deviation.
The comfort reward evaluates whether $a_t$ remains within smoothness limits:
$R_{\text{comfort}} = \mathbb{I}_\text{comf}\big(a_t ^ {\text{steer}}| < \beta_{\text{lat}} \text{ and } | a_t^{\text{acceleration}}| < \beta_{\text{long}}\big)$,
where $\mathbb{I}_\text{comf}$ is a binary indicator, \( \beta_{\text{lat}} \) and \( \beta_{\text{long}} \) are instruction-dependent lateral and longitudinal comfort thresholds, respectively.

\begin{figure}[!t]
  \centering
   \includegraphics[width=\linewidth]{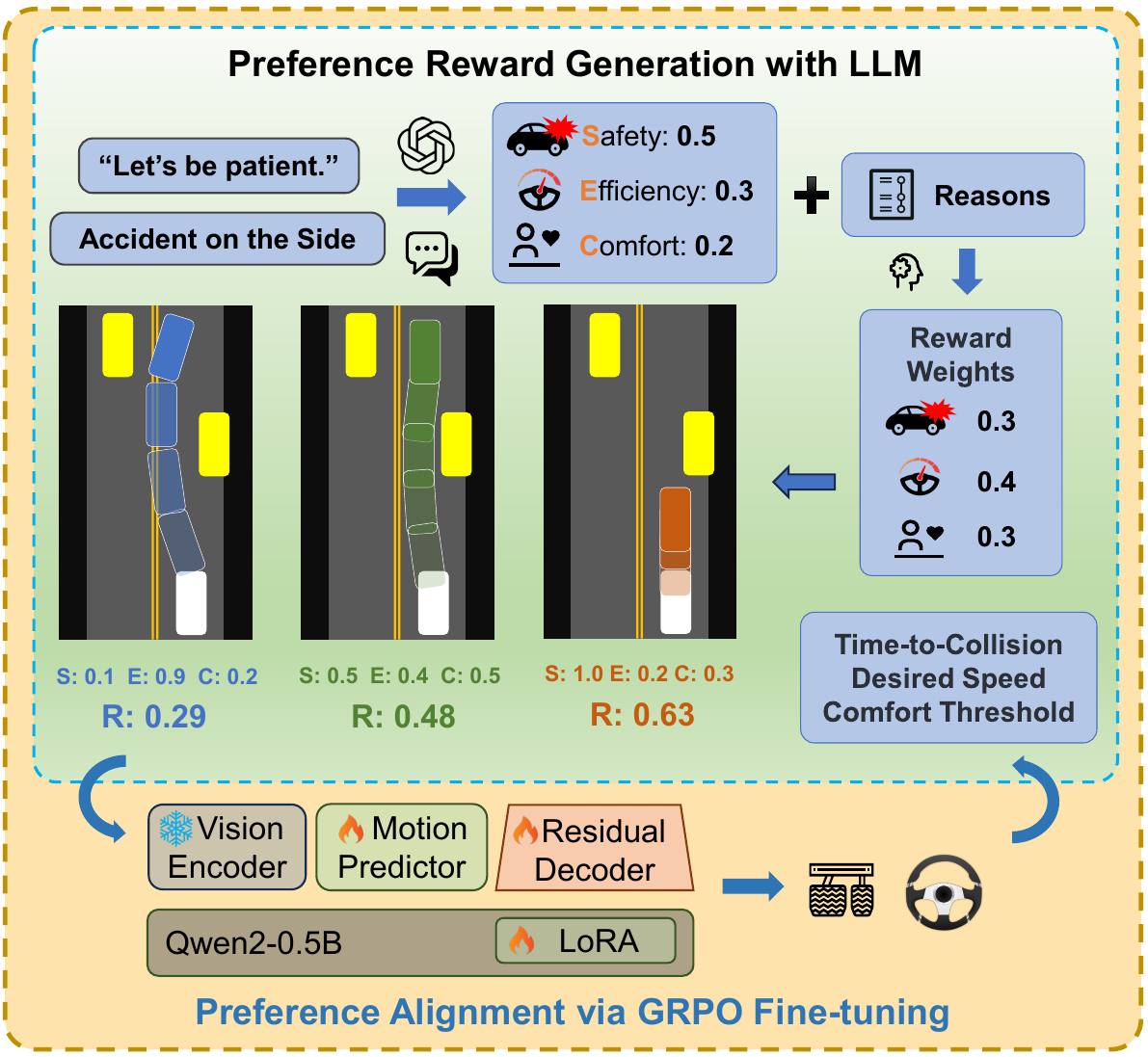}
   \vspace{-0.5cm}
   \caption{The fine-tuning process and reward generation for short-term instruction alignment.}
   \vspace{-0.3cm}
   \label{fig:grpo}
\end{figure}

To encourage personalized interaction with the user and adapt to real-time preferences, we map the reward parameters based on both the instruction and the driving context. Specifically, for each style command $I_t$, and scenario description, this mapping produces a set of weights $(w_s, w_e, w_c)$, a TTC threshold $\beta_\text{safety}$, a desired speed $v_{\text{pref}}$, and comfort thresholds $\beta_\text{lat}$, $\beta_\text{long}$. Instructions associated with an aggressive style correspond to a higher $w_e$ and $v_\text{pref}$ to prioritize efficiency over other metrics. This formulation provides a more distinct optimization objective for each style rather than relying on a generic combination.

To enable scalable and reliable weight adjustment, we employ a multi-stage approach illustrated in \cref{fig:grpo}. First, we leverage the reasoning capabilities of advanced LLMs (e.g., GPT-5) to infer reward weights and thresholds from the scenario description and the instruction with a style $S \in \{\text{Conservative}, \text{Neutral}, \text{Aggressive}\}$, and the inferred parameters are initialized within predefined upper bounds for each style to maintain balanced trade-offs among metrics. Second, the generated parameters are refined through expert review to ensure that the final reward functions reflect the intended command semantics while preserving safe driving behaviors across diverse traffic conditions.
This adaptation bridges the gap between subtle preferences in language and dynamic style rewards.

%% file: sec/6_experiments.tex
\section{Experiments}
\label{sec:experiments}

We conduct extensive closed-loop experiments to validate the effectiveness of the DMW framework in both user alignment and instruction preference adaptation. Our evaluation is designed to answer three key research questions: 

\begin{itemize} 
\item \textbf{RQ1: Long-term Driving Alignment.} Can our policy align with specific driving behavior when conditioned on learned user embeddings?
\item \textbf{RQ2: Short-term Instruction Adaptation.} How well does the policy align with various styles of real-time language commands under different scenarios? 
\item \textbf{RQ3: Driving Performance.} Does the introduction of personalization maintain or compromise driving performance, particularly regarding safety and success rate in complex scenarios? 
\end{itemize}

\subsection{Experimental Setup}
\textbf{Implementation Details.}
For user embedding learning, both the long-term preference encoder and the route processor are trained using the AdamW optimizer~\cite{loshchilovdecoupled} with a weight decay of 1e-3 and a learning rate of 1e-4. After convergence, the long-term preference encoder is frozen, and we fine-tune the full motion predictor and residual decoder while adopting the LoRA adapter~\cite{hu2022lora} for parameter-efficient adaptation of Qwen2-0.5B~\cite{yang2024qwen2}. The full model is trained on eight NVIDIA RTX A6000 GPUs with a per-GPU batch size of 8. For GRPO, we generate 4 responses per input for policy gradient updates.

\begin{table*}[t!]
\centering
\caption{\textbf{Driving metrics with and without style instructions, for each driver profile with a different long-term preference. Note that for Alignment Score (AS) and Ratings, we compute their value regardless of the style since it only measures long-term alignment.}}
\vspace{-0.2cm}
\footnotesize

% \scalebox{0.99}
{%
{\renewcommand{\arraystretch}{0.9}%
\resizebox{1.0\textwidth}{!}{
\begin{tabular}{l|c|c|>{\centering\arraybackslash}c
                    >{\centering\arraybackslash}c
                    >{\centering\arraybackslash}c
                    >{\centering\arraybackslash}c
                    >{\centering\arraybackslash}c
                    |>{\centering\arraybackslash}c
                    >{\centering\arraybackslash}c}
\toprule
\textbf{Driver} & \textbf{Scenario} & \textbf{Style} & DS & Speed & Efficiency & Acceleration & Headway & AS & Ratings\\
\midrule
% ---------------- Emergency Brake ----------------
D1 & \multirow{4}{*}{\shortstack{Emergency\\Brake}} 
  & \multirow{4}{*}{\shortstack{None \\ $\downarrow$ \\ Aggressive}} 
  & 95.10 → 84.02 & 8.06 → 9.58 & 150.94 → 167.12 & 6.31 → 6.39 & 18.94 → 17.11 & 1.00 & 8.8\\
D2 & & & 98.64 → 96.71 & 4.61 → 4.97 & 96.02 → 99.21 & 4.94 → 5.35 & 23.10 → 21.18 & 0.67 & 8.3\\
D3 & & & 97.88 → 95.93 & 5.39 → 6.34 & 109.83 → 121.06 & 5.63 → 6.16 & 21.61 → 19.22 & 1.00 & 8.0 \\
D4 & & & 94.02 → 93.15 & 7.76 → 8.35 & 145.22 → 159.56 & 6.03 → 6.92 & 21.03 → 18.19 & 0.67 & 7.8\\
\midrule
% ---------------- Merging ----------------
D1 & \multirow{4}{*}{Merging}
  & \multirow{4}{*}{\shortstack{None\\ $\downarrow$ \\ Conservative}}
  & 90.05 → 94.38 & 9.32 → 8.76 & 270.84 → 260.10 & 5.15 → 4.68 & 96.44 → 101.16 & 0.67 & 8.4\\
D2 & & & 86.94 → 91.72 & 6.37 → 5.76 & 196.82 → 175.31 & 4.29 → 4.16 & 118.92 → 121.03 & 1.00 & 8.1\\
D3 & & & 96.85 → 95.94 & 6.98 → 6.20 & 205.41 → 195.77 & 4.46 → 4.24 & 111.26 → 114.05 & 0.67 & 7.6\\
D4 & & & 97.31 → 97.88 & 8.78 → 8.63 & 261.52 → 259.52 & 5.01 → 4.91 & 88.76 → 100.93 & 1.00 & 8.2\\
\midrule
% ---------------- Overtaking ----------------
D1 & \multirow{4}{*}{Overtaking}
  & \multirow{4}{*}{\shortstack{None\\ $\downarrow$ \\ Neutral}}
  & 97.56 → 97.94 & 7.70 → 7.94 & 220.19 → 223.13 & 7.28 → 7.37 & 14.69 → 14.53 & 1.00 & 9.0\\
D2 & & & 98.23 → 98.41 & 5.50 → 5.56 & 167.40 → 167.80 & 5.65 → 5.78 & 22.23 → 22.04 & 1.00 & 8.6\\
D3 & & & 97.14 → 96.62 & 6.34 → 6.13 & 191.02 → 183.84 & 6.18 → 6.01 & 20.79 → 22.19 & 0.67 & 7.5\\
D4 & & & 80.61 → 82.19 & 9.63 → 9.51 & 271.98 → 265.84 & 7.11 → 6.82 & 17.81 → 18.03 & 1.00 & 8.1\\
\midrule
% ---------------- Traffic Sign ----------------
D1 & \multirow{4}{*}{\shortstack{Traffic\\Sign}}
  & \multirow{4}{*}{\shortstack{None\\ $\downarrow$ \\ Conservative}}
  & 90.82 → 96.67 & 9.08 → 8.49 & 408.74 → 401.28 & 7.12 → 6.37 & 31.44 → 32.52 & 1.00 & 8.7\\
D2 & & & 97.92 → 98.35 & 5.79 → 4.91 & 193.26 → 183.11 & 5.72 → 5.51 & 35.02 → 36.91 & 1.00 & 8.2\\
D3 & & & 89.31 → 92.02 & 6.88 → 5.37 & 237.05 → 231.46 & 5.96 → 5.91 & 34.54 → 36.51 & 1.00 & 8.1\\
D4 & & & 95.91 → 97.32 & 8.91 → 6.74 & 418.02 → 379.68 & 6.54 → 5.81 & 31.19 → 31.77 & 0.67 & 7.9\\
\bottomrule
\end{tabular}%
}
}%
}%
\vspace{-0.3cm}
\label{tab:user_study}
% \vspace{-0.4cm}
\end{table*}

\noindent\textbf{Scenarios and Baselines.} We evaluate closed-loop performance using Bench2Drive~\cite{jia2024bench2drive}. Unlike the original evaluation, which continues the episode after a collision, we terminate the route immediately once a collision occurs to emphasize the safety-critical assessment. To validate the effect of personalization, we compare DMW with SimLingo~\cite{renz2025simlingo} under various styles of instruction, focusing on whether the model can effectively adjust its driving tendency according to user-specific preferences. 
Additionally, to strengthen comparisons with prior personalization-focused work, we implement a StyleDrive-like \cite{hao2026styledrive} baseline by mapping each instruction to a style condition (i.e., Aggressive, Neutral, Conservative) and injecting it into the policy. Meanwhile, we adopt MORL-PD \cite{surmann2025multi} as a baseline since it conditions the policy on a preference vector (speed/comfort) at runtime to modulate behavior. By deriving a per-user preference vector from driving metrics of test drivers and conditioning a multi-objective RL (MORL) policy, we compare it with DMW for long-term preference alignment.

\noindent\textbf{Evaluation Metrics.} We employ metrics that assess both driving performance and stylistic alignment. The Driving Score (DS), Success Rate (SR), Efficiency (Effic.), and Comfort are adopted from Bench2Drive \cite{jia2024bench2drive}. To quantify the personalization of the model, we measure the mean value of driving metrics, including average Speed in $\SI{}{m/s}$, Acceleration (Acce.) in $\SI{}{m/s^2}$, Lane Change Counts (LC), Headway in $\SI{}{m}$, and Travel Time (TT) in $\SI{}{s}$.
We introduce the Alignment Score (AS) for user studies to evaluate how well the policy aligns with individual driving preferences.

\noindent\textbf{User Study for Long-Term Preference Alignment.}
To test whether the driving behaviors of DMW align with each driver's preference. We introduce an Alignment Score (AS) by first clustering all drivers using their historical logs, then generating roll-outs condition on each driver's profile, and finally calculating the accuracy across test routes where roll-out is correctly classified into their target cluster. 
Additionally, we recruit ten evaluators and ask them to rate the similarity of the driver's own logs and corresponding model roll-outs on a 1-10 scale.
To test the zero-shot generalizability of DMW, we perform preference alignment on 25 drivers from PDD and evaluate the behavior alignment on both 25 in-distribution (ID) and 5 out-of-distribution (OOD) drivers. For each scenario type, we choose three test routes.

\subsection{Main Results}
\textbf{Adaptation to Style Instruction.}
Table \ref{tab:bench2drive} presents the closed-loop evaluation results on the Bench2Drive. 
The results show that fine-tuning alone already improves both SR and DS, likely due to the newly introduced safety reward.
Furthermore, when conditioned on conservative instructions, DMW achieves the greatest gains in DS and SR while maintaining comparable efficiency, reflecting improved reliability in cautious driving.
Under the aggressive instructions, our policy yields a substantial efficiency gain of 18.77\% compared to the 3.70\% in SimLingo, and 6.00\% in StyleDrive, with only a 3.89\% reduction in DS relative to the conservative condition. DMW significantly outperforms SimLingo in style adaptation, which highlights that DMW accurately captures the intended behaviors expressed by the instructions while preserving overall performance.
Meanwhile, StyleDrive induces smaller metric shifts and achieves a lower DS than DMW, likely due to its fixed style condition, which further highlights the benefit of our style-sensitive adaptation.
The consistent reduction in acceleration and lane-change counts under conservative ones further validates that DMW maintains distinct stylistic behaviors. 

\begin{figure*}[!t]
  \centering
   \includegraphics[width=\textwidth]{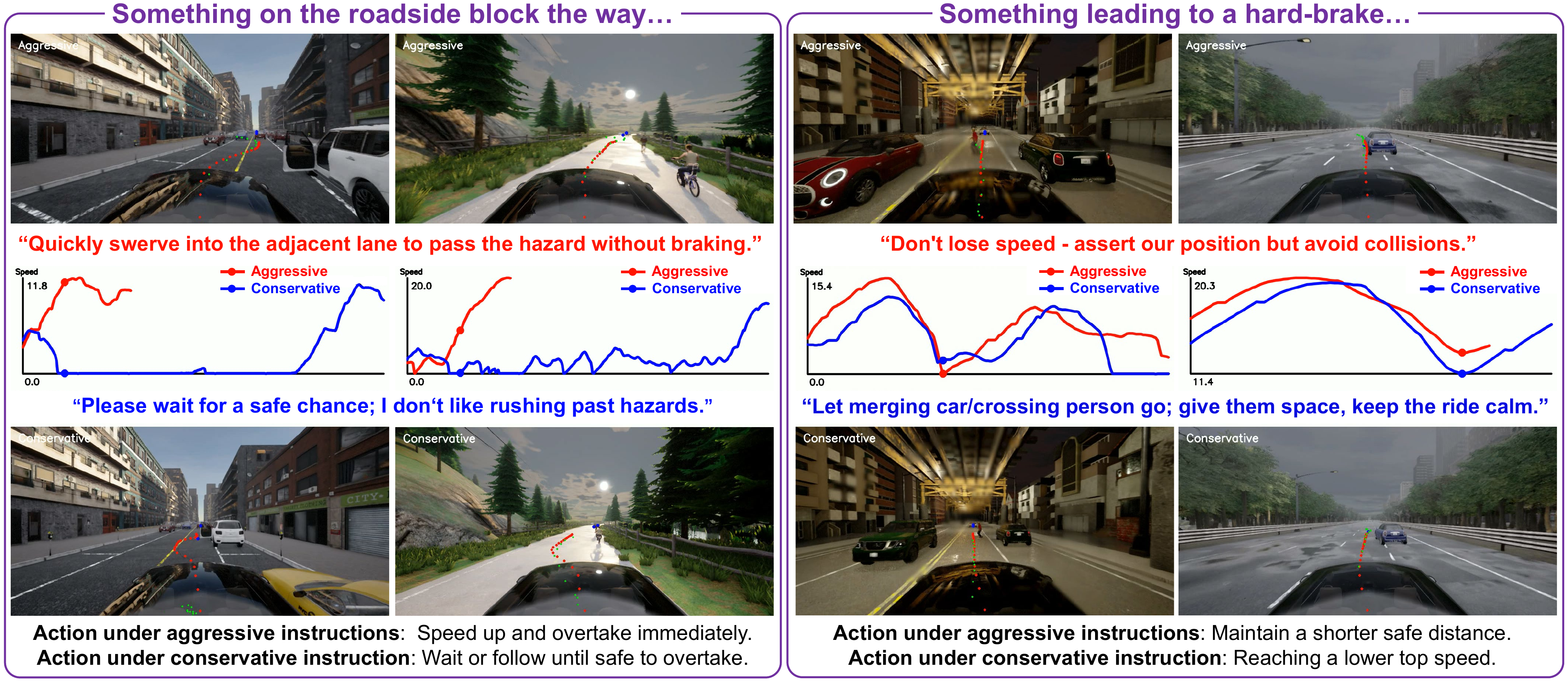}
   \vspace{-0.5cm}
   \caption{Driving preference under aggressive and conservative instructions. \textcolor{red}{Red waypoints} denote distance parametrized (every \SI{1}{m}) navigation path and \textcolor{green}{green waypoints} denote time parametrized (every \SI{0.25}{s}) trajectory.}
   \vspace{-0.3cm}
   \label{fig:visualization}
\end{figure*}

To further illustrate the behavioral diversity induced by different instructions, \cref{fig:visualization} visualizes how our policy reacts under aggressive and conservative instructions in safety-critical scenarios. When facing a road blockage, the aggressive instruction prefers an immediate overtake (red speed curve). Conversely, the conservative command causes the agent to decelerate and either stop or maintain a low following speed until the opposite lane is clear (blue curve). This behavioral divergence arises from the style-aware weighting in our fine-tuning reward, which dynamically adjusts the balance between safety and efficiency according to the given instruction. 
In hard-braking scenarios, the agent tends to maintain a tighter following distance when objects appear ahead or merge into the lane under aggressive instructions, while the agent slows down and yields more noticeably when conditioned on conservative ones. This distinction is largely governed by the style-aware TTC threshold, which is adapted to be more permissive for aggressive ones and more restrictive for conservative ones, directly tuning the model's safety margins. The complete driving clips are provided in the supplementary materials. 

\noindent\textbf{Long-term Preference Alignment.}
We report metrics and user study results for two seen drivers (D1, D2) and two unseen drivers (D3, D4).
Table \ref{tab:user_study} shows that the policy conditioned on each profile exhibits consistent motion statistics across different scenarios. For instance, Drivers 1 and 4 exhibit higher efficiency and acceleration, reflecting a more aggressive driving preference, whereas Drivers 2 and 3 show lower acceleration and larger headway across all categories, indicating a preference that prioritizes safety and comfort. Furthermore, Driver 2 shows the most cautious style with the lowest acceleration, while Driver 3, though still conservative, exhibits slightly higher speeds and greater acceleration, suggesting a willingness to accelerate more readily in scenarios such as Overtaking and Traffic Sign. While each driver's driving metrics vary across scenarios, the driving behavior remains distinct among the drivers, suggesting that the policy aligns with individual drivers' long-term driving preferences across different scenarios. 
Furthermore, Table~\ref{tab:user_study_compare} shows higher AS and ratings for DMW, indicating that more roll-outs are classified into the target cluster where their driver belongs, and evaluators can consistently recognize driving behaviors that reflect the corresponding driver's driving habits. While MORL-PD also induces style shifts via its preference-weight conditioning, it achieves a lower alignment, especially on unseen drivers, which suggests that our user embedding better captures semantic context from the profile, leading to stronger generalization.

In addition, consistent with Table \ref{tab:bench2drive}, language style instructions can provide an additional adaptation, shifting the driving towards the short-term preference.

\begin{table}[t!]
\centering
\caption{Long-term preference alignment. D: Driver ID.}
\vspace{-0.2cm}
\footnotesize
\setlength{\tabcolsep}{3.6pt}
{\renewcommand{\arraystretch}{0.8}
\resizebox{0.95\linewidth}{!}{%
\begin{tabular}{l|cc|cc|cc|cc}
\toprule
& \multicolumn{4}{c|}{\textbf{Alignment Score}} & \multicolumn{4}{c}{\textbf{Average Ratings}} \\
\cmidrule(lr){2-5}\cmidrule(lr){6-9}
\multirow{2}{*}{\textbf{Methods}}
& \multicolumn{2}{c|}{\textbf{ID}} & \multicolumn{2}{c|}{\textbf{OOD}}
& \multicolumn{2}{c|}{\textbf{ID}} & \multicolumn{2}{c}{\textbf{OOD}} \\
\cmidrule(lr){2-3}\cmidrule(lr){4-5}\cmidrule(lr){6-7}\cmidrule(lr){8-9}
& D1 & D2 & D3 & D4
& D1 & D2 & D3 & D4 \\
\midrule
MORL-PD~\cite{surmann2025multi} &
0.42 & 0.58 & 0.25 & 0.33 &
5.1 & 6.2 & 3.9 & 3.5 \\
\textbf{DMW} &
0.92 & 0.92 & 0.83 & 0.83 &
8.7 & 8.3 & 7.8 & 8.0 \\
\bottomrule
\end{tabular}%
}}
\vspace{-0.4cm}
\label{tab:user_study_compare}
\end{table}

\subsection{Ablation Study}
\textbf{Adaptive Average Pooling.}
We ablate the masked Adaptive Average Pooling (AAP) module used in the preference encoder, which compresses textual profile features into a fixed-length sequence aligned with the temporal features of the route processor. Replacing this module with an unmasked global mean reduces the expressiveness of user embeddings. As shown in Table \ref{tab:user_study_ablation}, driving policy conditioned on the user embeddings from the preference encoder with AAP yields a higher diversity of motion statistics across different user profiles. We also observe a higher AS, demonstrating that the model can more consistently reproduce each driver’s behavioral traits. We attribute the improvement to its ability to preserve semantically important embeddings while mitigating the bias from irrelevant segments. 
We also observe that in denser interaction scenarios, the policy may shift toward conservative behavior for safety, which can reduce style expressiveness for neutral drivers.

\begin{table}[t!]
\centering
\caption{\textbf{Averaged driving metrics using long-term preference encoder with or without Adaptive Average Pooling (AAP).}}
\vspace{-0.2cm}
\footnotesize
\setlength{\tabcolsep}{2pt} % adjust column spacing if needed
{\renewcommand{\arraystretch}{0.8} % adjust row spacing
\begin{tabular}{l|c|ccccc|cc}
\toprule
\textbf{Driver} & \textbf{Encoder} & DS & Speed & Effic. & Acce.  & Headway & AS & Ratings \\ 
\midrule
D1 & \multirow{4}{*}{w/o AAP} & 82.74  & 4.52 & 161.38 & 5.84 & 50.11 & 0.67 & 6.1\\
D2 & & 88.02 & 4.21 & 163.07 & 5.88 & 47.56 & 0.58 & 5.9\\
D3 & & 87.65 & 4.49 & 133.12 & 5.81 & 41.63 & 0.25 & 4.6\\
D4 & & 80.13  & 4.46 & 172.92 & 5.79 & 38.94 & 0.50 & 5.3\\
\midrule
D1 & \multirow{4}{*}{w/ AAP} & 93.38 & 8.54 & 262.68 & 6.46 & 40.38 & 0.92 & 8.7\\
D2 & & 95.43 & 5.57 & 163.38 & 5.15 & 49.82 & 0.92 & 8.3\\
D3 & & 95.30 & 6.40 & 185.83 & 5.56 & 47.05 & 0.83 & 7.8\\
D4 & & 91.96 & 8.77 & 274.19 & 6.17 & 39.70 & 0.83 & 8.0\\
\bottomrule
\end{tabular}}
\vspace{-0.3cm}
\label{tab:user_study_ablation}
% \vspace{-0.4cm}
\end{table}

\noindent\textbf{Style-aware Reward Adaptation.} We examine the effect of fixed versus style-aware adaptive reward weights on driving performance and behavioral diversity. The results are in Table \ref{tab:bench2drive}. When the reward components are fixed (i.e., $w_s = 0.35$, $w_e = 0.35$, $w_c = 0.30$), only the style-dependent TTC and comfort thresholds remain to induce stylistic variation. DMW-Vanilla achieves higher overall DS and SR, indicating effective optimization of generic objectives. However, the inability of dynamic adaptation to efficiency and comfort weights causes aggressive and conservative instructions to exhibit similar driving, revealing a reduced sensitivity to style.
In contrast, DMW adaptively adjusts reward weights and thresholds based on both context and instruction style, maintaining driving performance while producing clear behavioral distinctions. Aggressive instructions lead to higher speeds and shorter headways, whereas conservative ones favor smoother control and larger safety margins. This demonstrates the necessity of adaptive weighting to achieve both reliable and personalized driving behaviors.

%% file: sec/7_conclusion.tex
\section{Conclusion}
\label{sec:conclusions}
We introduced DMW, a personalized VLA driving framework that aligns long-term driver preferences and adapts to short-term style instructions. 
Through learning user embeddings from the curated Personalized Driving Dataset and conditioning on the end-to-end driving policy, our method captures and aligns with individual drivers' behaviors. 
Extensive experiments on Bench2Drive demonstrate that DMW achieves more distinct adaptation than existing baselines. 
User studies further highlight the potential of personalized VLA systems to bridge the gap toward human-centered autonomous driving, enabling vehicles that adapt their behavior to individual user preferences.
Our approach is currently validated within the CARLA simulator. Future work will focus on bridging the sim-to-real gap by deploying and evaluating the DMW framework on real vehicles. 
We also plan to expand our dataset to encompass a wider diversity of driving behaviors to enable more robust and generalizable preference alignment.

%% file: sec/X_suppl.tex
\clearpage
\setcounter{page}{1}
\maketitlesupplementary

\section{Additional Quantitative Results}
To further validate whether DMW produces behaviors consistent with human expectations, we report the detailed user study results on both long-term preference alignment and short-term adaptation to style instructions.

\subsection{Long-term Preference Alignment}
Building on the results across scenario types in Table \ref{tab:user_study}, the averaged metrics in Table \ref{tab:user_study_full} further show clear differences in driver-specific behavior. Drivers with higher average speeds exhibit consistent behavior across scenarios, whereas more cautious drivers tend to maintain larger headways and lower accelerations. These trends are consistent with the cross-scenario patterns observed in Table \ref{tab:user_study}, suggesting that the policy aligns with persistent behavioral traits beyond individual routes.

\subsection{Adaptation on Style Instruction}
Beyond long-term preferences, users may also express short-term intentions via style instructions depending on situational context. To complement objective metrics in Table \ref{tab:bench2drive}, and further evaluate whether the policy adapts to these instructions, we conduct a user study in which human evaluators rate trajectories generated under different style prompts for the same scenario. Each evaluator is shown short video clips rendered under \textit{conservative}, \textit{neutral}, and \textit{aggressive} instructions and asked to judge whether the resulting behavior matches the intended style.

Evaluators score each trajectory on a 0-10 scale according to three criteria:  
(i) how well the behavior follows the instruction,  
(ii) efficiency, comfort, and smoothness of drive, and  
(iii) perceived safety.  
As shown in Table \ref{tab:user_study_instruction}, across four representative scenario types, both StyleDrive~\cite{hao2026styledrive} and DMW outperform SimLingo baseline~\cite{renz2025simlingo}, indicating the benefit of style-aware driving adaptation. StyleDrive~\cite{hao2026styledrive} demonstrates improved alignment with short-term style instructions compared to SimLingo, but still falls short of DMW. In contrast, DMW consistently achieves the highest ratings across all styles and scenarios. Evaluators consistently observe that under aggressive instructions, DMW produces higher speeds, shorter following distances, and more decisive accelerations, while conservative instructions yield smoother control profiles and larger safety margins. These results confirm that the proposed policy not only responds to explicit style instructions, but does so in a more sensitive manner.

\begin{table}[t!]
\centering
\small
\setlength{\tabcolsep}{3pt}
\caption{Driving metrics across all scenario types.}
\vspace{-0.2cm}
\resizebox{1.0\linewidth}{!}{
\begin{tabular}{l|ccccc|cc}
\toprule
\textbf{Driver} & DS & Speed & Effic. & Acce. & Headway & AS & Ratings \\
\midrule
D1  & 93.38 & 8.54 & 262.68 & 6.46 & 40.38 & 0.92 & 8.7\\
D2  & 95.43 & 5.57 & 163.38 & 5.15 & 49.82 & 0.92 & 8.3\\
D3  & 95.30 & 6.40 & 185.83 & 5.56 & 47.05 & 0.83 & 7.8\\
D4  & 91.96 & 8.77 & 274.19 & 6.17 & 39.70 & 0.83 & 8.0\\
D5  & 92.58 & 7.48 & 229.93 & 5.58 & 39.08 & 0.83 & 7.9 \\
D6  & 97.71 & 6.16 & 190.74 & 6.08 & 42.41 & 0.75 & 7.4 \\
D7  & 96.11 & 7.16 & 218.25 & 6.36 & 40.12 & 0.83 & 7.9 \\
D8  & 95.28 & 5.82 & 188.27 & 5.54 & 43.95 & 0.92 & 8.2 \\
D9  & 96.16 & 5.60 & 178.97 & 5.84 & 45.07 & 0.67 & 7.0 \\
D10 & 94.20 & 8.38 & 258.28 & 6.89 & 40.26 & 1.00 & 8.6 \\
\bottomrule
\end{tabular}}
\vspace{-0.3cm}
\label{tab:user_study_full}
\end{table}

\begin{table}[t!]
\centering
\small
\setlength{\tabcolsep}{4pt}
\renewcommand{\arraystretch}{1.0}
\caption{User study ratings (0-10) evaluating how well trajectories match intended instructions. Five evaluators (E1-E5) rate trajectories from SimLingo~\cite{renz2025simlingo}, StyleDrive~\cite{hao2026styledrive}, and \textbf{DMW}.}
\label{tab:user_study_instruction}
\resizebox{1.0\columnwidth}{!}{
\begin{tabular}{l|c|c|ccccc}
\toprule
\textbf{Scenario} & \textbf{Model} & \textbf{Style} 
& \textbf{E1} & \textbf{E2} & \textbf{E3} & \textbf{E4} & \textbf{E5} \\
\midrule

% ---------------- Emergency Brake ----------------
\multirow{9}{*}{\shortstack{Emergency\\Brake}}
  & \multirow{3}{*}{SimLingo~\cite{renz2025simlingo}}
    & Conservative & 7.4 & 6.7 & 6.4 & 7.3 & 6.5 \\
  & & Neutral      & 7.0 & 7.3 & 7.5 & 7.0 & 7.1 \\
  & & Aggressive   & 6.5 & 7.6 & 7.2 & 6.6 & 7.4 \\
  \cmidrule(lr){2-8}
  & \multirow{3}{*}{StyleDrive~\cite{hao2026styledrive}}
    & Conservative & 8.2 & 7.5 & 7.3 & 8.0 & 7.2 \\
  & & Neutral      & 7.8 & 8.0 & 8.2 & 7.6 & 7.9 \\
  & & Aggressive   & 7.2 & 8.4 & 7.9 & 7.1 & 8.2 \\
  \cmidrule(lr){2-8}
  & \multirow{3}{*}{\textbf{DMW}}
    & Conservative & 9.0 & 8.1 & 8.0 & 8.8 & 7.9 \\
  & & Neutral      & 8.4 & 8.6 & 9.1 & 8.1 & 8.5 \\
  & & Aggressive   & 7.8 & 9.2 & 8.3 & 7.6 & 9.0 \\
\midrule

% ---------------- Merging ----------------
\multirow{9}{*}{Merging}
  & \multirow{3}{*}{SimLingo~\cite{renz2025simlingo}}
    & Conservative & 7.2 & 6.4 & 6.3 & 7.2 & 6.3 \\
  & & Neutral      & 6.8 & 7.0 & 7.3 & 6.7 & 6.8 \\
  & & Aggressive   & 6.2 & 7.5 & 6.9 & 6.1 & 7.3 \\
  \cmidrule(lr){2-8}
  & \multirow{3}{*}{StyleDrive~\cite{hao2026styledrive}}
    & Conservative & 8.0 & 7.3 & 7.2 & 7.9 & 7.1 \\
  & & Neutral      & 7.6 & 7.8 & 8.1 & 7.4 & 7.7 \\
  & & Aggressive   & 7.0 & 8.3 & 7.6 & 6.9 & 8.1 \\
  \cmidrule(lr){2-8}
  & \multirow{3}{*}{\textbf{DMW}}
    & Conservative & 8.9 & 7.9 & 7.7 & 8.7 & 7.6 \\
  & & Neutral      & 8.2 & 8.4 & 9.0 & 8.0 & 8.3 \\
  & & Aggressive   & 7.5 & 9.1 & 8.1 & 7.4 & 8.9 \\
\midrule

% ---------------- Overtaking ----------------
\multirow{9}{*}{Overtaking}
  & \multirow{3}{*}{SimLingo~\cite{renz2025simlingo}}
    & Conservative & 7.3 & 6.3 & 6.2 & 7.2 & 6.2 \\
  & & Neutral      & 6.9 & 7.1 & 7.4 & 6.8 & 6.9 \\
  & & Aggressive   & 6.1 & 7.5 & 7.0 & 6.0 & 7.3 \\
  \cmidrule(lr){2-8}
  & \multirow{3}{*}{StyleDrive~\cite{hao2026styledrive}}
    & Conservative & 8.1 & 7.2 & 7.1 & 8.0 & 7.0 \\
  & & Neutral      & 7.7 & 7.9 & 8.2 & 7.5 & 7.8 \\
  & & Aggressive   & 6.9 & 8.4 & 7.7 & 6.8 & 8.2 \\
  \cmidrule(lr){2-8}
  & \multirow{3}{*}{\textbf{DMW}}
    & Conservative & 9.1 & 7.7 & 7.5 & 8.8 & 7.6 \\
  & & Neutral      & 8.3 & 8.5 & 9.0 & 8.0 & 8.4 \\
  & & Aggressive   & 7.4 & 9.3 & 8.2 & 7.2 & 9.2 \\
\midrule

% ---------------- Traffic Sign ----------------
\multirow{9}{*}{\shortstack{Traffic\\Sign}}
  & \multirow{3}{*}{SimLingo~\cite{renz2025simlingo}}
    & Conservative & 7.4 & 7.0 & 6.9 & 7.3 & 6.8 \\
  & & Neutral      & 7.1 & 7.3 & 7.5 & 7.0 & 7.1 \\
  & & Aggressive   & 6.8 & 7.6 & 7.2 & 6.7 & 7.4 \\
  \cmidrule(lr){2-8}
  & \multirow{3}{*}{StyleDrive~\cite{hao2026styledrive}}
    & Conservative & 8.3 & 7.8 & 7.6 & 8.1 & 7.7 \\
  & & Neutral      & 7.9 & 8.1 & 8.3 & 7.7 & 8.0 \\
  & & Aggressive   & 7.4 & 8.6 & 8.0 & 7.2 & 8.4 \\
  \cmidrule(lr){2-8}
  & \multirow{3}{*}{\textbf{DMW}}
    & Conservative & 8.9 & 8.3 & 8.1 & 8.6 & 8.2 \\
  & & Neutral      & 8.5 & 8.7 & 8.8 & 8.2 & 8.6 \\
  & & Aggressive   & 8.0 & 9.0 & 8.4 & 7.8 & 8.9 \\
\bottomrule
\end{tabular}
}
\vspace{-0.3cm}
\end{table}

\subsection{Additional Qualitative Results}

\begin{figure}[t]
    \centering
    \begin{subfigure}{\linewidth}
        \centering
        \includegraphics[width=\linewidth]{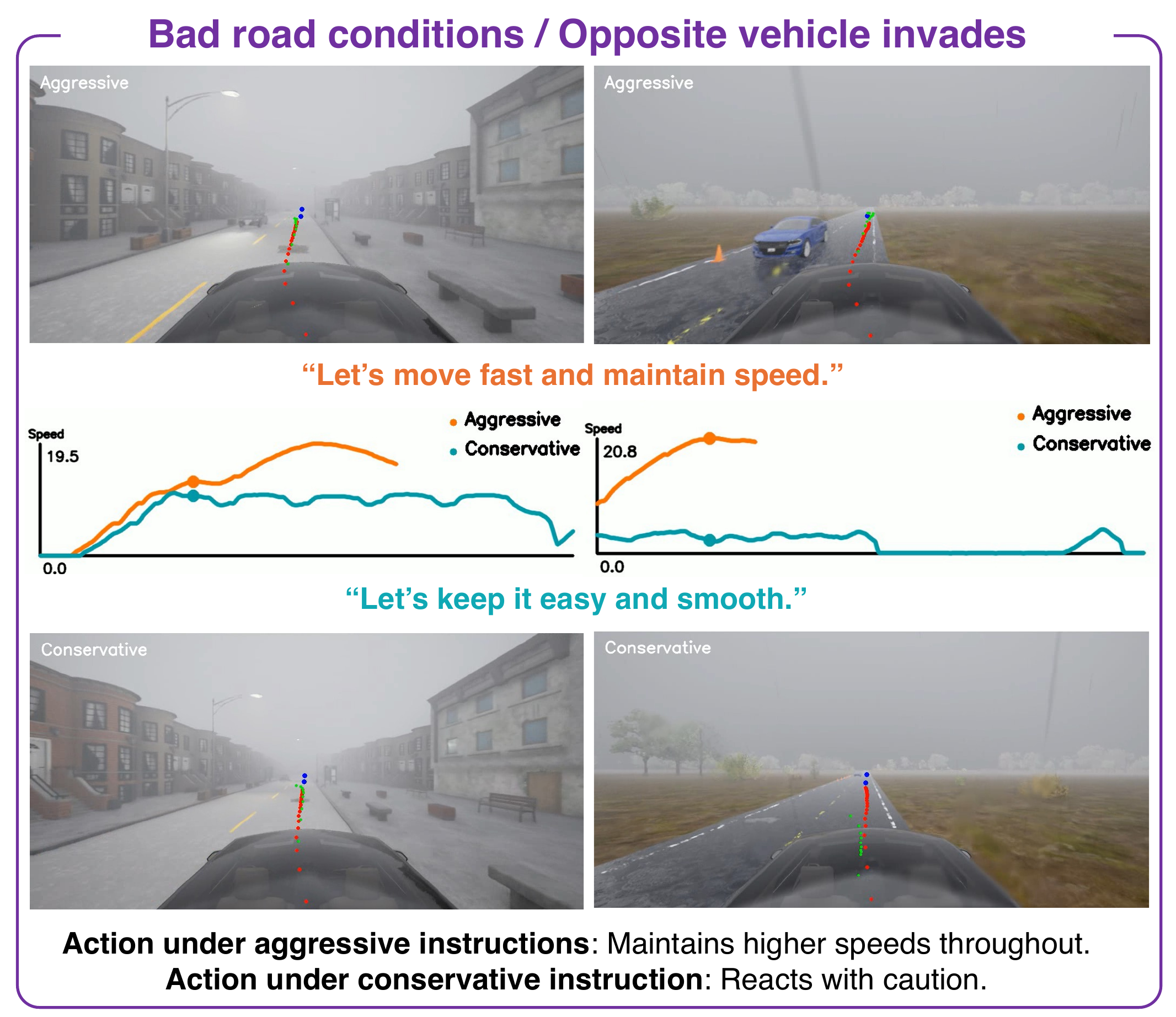}
    \end{subfigure}
    \vspace{0.3cm}
    \begin{subfigure}{\linewidth}
        \centering
        \includegraphics[width=\linewidth]{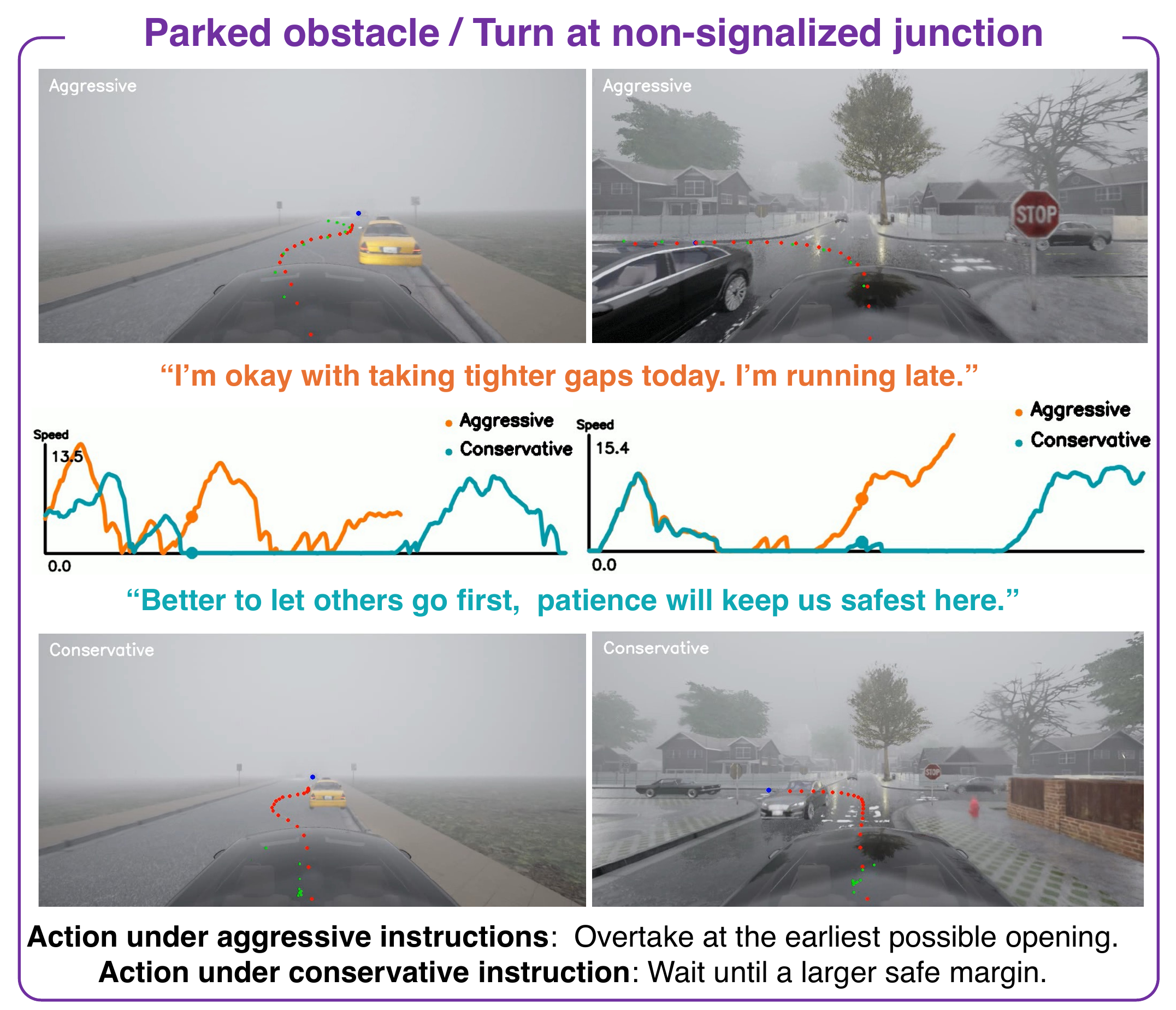}
    \end{subfigure}
    \vspace{-0.6cm}
    \caption{Driving preference under aggressive and conservative instructions. \textcolor{red}{Red waypoints} denote distance parametrized (every \SI{1}{m}) navigation path and \textcolor{green}{green waypoints} denote time parametrized (every \SI{0.25}{s}) trajectory.}
    \vspace{-0.3cm}
    \label{fig:suppl_visualization}
\end{figure}

\cref{fig:suppl_visualization} visualizes how DMW responds to aggressive and conservative instructions across safety-critical scenarios. In the lost-of-control scenario, where the ego-vehicle risks losing control due to the bad road conditions, the aggressive instruction prioritizes efficiency: it maintains a higher speed and attempts to pass the unstable area quickly. In contrast, the conservative instruction leads the agent to maintain a smoother trajectory, and preserve vehicle stability. This divergence highlights how the policy adapts its balance between efficiency and comfort based on the given instruction.

A similar pattern emerges in the oncoming-vehicle intrusion scenario. When another vehicle invades the ego lane, the aggressive instruction causes the agent to accelerate and execute a decisive rightward maneuver at relatively high speed. Meanwhile, the conservative instruction prompts early caution: the agent reduces speed, yields space sooner, and performs a safer avoidance. 

In another scenario involving a parked vehicle blocking the lane, the agent must decide when to safely overtake. Under an aggressive instruction, corresponding to personal requirements such as “I’m in a hurry” or “I’m running late”. The policy seeks the earliest viable gap and initiates the overtake quickly to reduce waiting time. In contrast, under a conservative instruction, the agent remains patient, yielding until the oncoming lane is fully clear, especially under low-visibility conditions.

In the scenario where the agent needs to make a left turn at an unsignalized junction, under the aggressive instruction, the agent initiates the turn earlier once it identifies a tighter and feasible opening, minimizing delay. In contrast, the conservative instruction causes the agent to wait patiently for a safer gap before turning. This cautious behavior reflects the emphasis on safety and low-risk in complex intersection negotiations. Together, these examples illustrate DMW’s ability to adapt in real-time according to short-term style instruction.

\section{Personalized Driving Dataset}
We provide a more detailed description of the collected personalized driving datasets from thirty real drivers. 

\subsection{Scenarios}
In Town 12, we collect twenty routes that cover a diverse set of representative driving scenarios under varying weather and illumination conditions. The scenarios are summarized below (descriptions are taken from \url{https://leaderboard.carla.org/scenarios/}):

\begin{itemize}
    \item \textbf{Accident / ParkedObstacle / ConstructionObstacle}: An obstacle (e.g., a construction zone, an accident, or a parked vehicle) is blocking the ego lane. The ego vehicle must change lanes into traffic moving in the same direction to bypass the obstacle.
    \item \textbf{SignalizedJunctionLeftTurn / NonSignalizedJunctionLeftTurn}: The ego vehicle performs an unprotected left turn at an intersection (can occur at both signalized and unsignalized intersections).
    \item \textbf{CrossingBicycleFlow}: The ego vehicle must execute a turn at an intersection while yielding to bicycles crossing perpendicular to its path.
    \item \textbf{StaticCutIn}: Another vehicle cuts into the ego lane from a queue of stationary traffic. It must decelerate, brake, or change lanes to avoid a collision.
    \item \textbf{NonSignalizedJunctionRightTurn / SignalizedJunctionRightTurn / VanillaNonSignalizedTurn}: The ego vehicle makes a right turn at an intersection while yielding to crossing traffic.
    \item \textbf{InterurbanActorFlow}: The ego vehicle leaves the interurban road by turning left, crossing a fast traffic flow.
    \item \textbf{BlockedIntersection}: While performing a maneuver, the ego vehicle encounters a stopped vehicle on the road and must perform an emergency brake or an avoidance maneuver.
    \item \textbf{HazardAtSideLane}: A slow-moving hazard (e.g., bicycle) partially obstructs the ego vehicle’s lane. The ego vehicle must either brake or carefully bypass the hazard (bypassing on the lane with traffic in the same direction).
    \item \textbf{ParkingCutIn}: A parked vehicle exits a parallel parking space into the ego vehicle’s path. The ego vehicle must slow down to allow the parked vehicle to merge into traffic.
    \item \textbf{VehicleOpensDoorTwoWays}: The ego vehicle needs to avoid a parked vehicle with its door opening into the lane.
    \item \textbf{DynamicObjectCrossing}: A pedestrian suddenly emerges from behind a parked vehicle and enters the lane. The ego vehicle must brake or take evasive action to avoid hitting the pedestrian.
    \item \textbf{EnterActorFlow}: A flow of cars runs a red light in front of the ego when it enters the junction, forcing it to react (interrupting the flow or merging into the flow). These vehicles are ’special’ ones, such as police cars, ambulances, or firetrucks.
    \item \textbf{HighwayExit}: The ego vehicle must cross a lane of moving traffic to exit the highway at an off-ramp.
    \item \textbf{ControlLoss}: The ego vehicle loses control due to bad conditions on the road and it must recover, coming back to its original lane.
    \item \textbf{MergerIntoSlowTraffic}: The ego-vehicle merge into a slow traffic on the off-ramp when exiting the highway.
\end{itemize}

\subsection{Auxiliary Information}
To enable reliable and interpretable driving preference analysis, we extract environmental information and motion statistics from driving logs. Concretely, we record data at \SI{5}{Hz}, including:

\begin{itemize}
    \item \textbf{Camera Sensor.}
    A forward-facing RGB camera with a resolution of \SI{1024}{\times}\SI{512} and a wide \ang{110} field of view serves as the primary visual sensor.
    
    \item \textbf{Ego-state and Control Signals.}  
    We store full ego-vehicle kinematics, including linear acceleration, angular velocity, speed, and the world-frame pose (\texttt{location}, \texttt{rotation}).  
    Human control commands, including throttle, brake, steering, gear, hand-brake status, and reverse flag, together with the speed limit and lane-level attributes such as lane ID, lane type, lane width, and whether the ego vehicle is currently inside a junction.

    \item \textbf{Surrounding Agents.}  
    We store a detailed description of the leading vehicle (\texttt{front\_vehicle\_info}: ID, type, 3D position, velocity, speed, and color), along with all nearby dynamic agents in \texttt{other\_vehicles} and \texttt{walkers}. For each surrounding actor, we log its transform, velocity, bounding-box extent, and other metadata. 
    Nearby traffic lights and stop signs are also recorded, capturing both their spatial relation to the ego and whether they are currently influencing the ego vehicle.

    \item \textbf{Expert Supervision.}  
    Each timestep is paired with privileged control from the PDM-Lite expert, including \texttt{throttle}, \texttt{brake}, \texttt{steer}, and the corresponding \texttt{target\_speed}. 

    \item \textbf{Route Geometry.}
    We additionally provide route information in the form of two upcoming waypoints along the global route, transformed into the ego frame as \texttt{target\_point} and \texttt{target\_point\_next}.

\end{itemize}

All actor-level annotations are saved as JSON files synchronized with the image index.